\documentclass[letterpaper, 10 pt, conference]{ieeeconf}  % Comment this line out if you need a4paper
\usepackage{xcolor}
\usepackage{tabularx}
\usepackage{adjustbox}
\usepackage{cite}
\usepackage{graphicx}
\usepackage{eurosym}
\usepackage{caption}
\usepackage{amsmath}
\usepackage{subcaption}
\usepackage{array}
\usepackage{booktabs}
\usepackage{seqsplit}
\newcolumntype{L}[1]{>{\raggedright\arraybackslash}p{#1}}
\usepackage{tikz}

\usepackage{placeins}
\usepackage{float}
\usepackage{stfloats}

\IEEEoverridecommandlockouts                              % This command is only needed if 
\title{\LARGE \bf
PUPPET: Physical-Unified Platform for Presence, Embodiment, and Teleoperation}

\author{Mahika Vats, Milan Groshev, Carlos Gomez Cubero, Eduardo Castell{\'{o}} Ferrer\\
\normalsize{CyPhy Life, Robotics \& AI Lab, School of Science \& Technology, IE University, Spain}}

\begin{document}

\maketitle
\thispagestyle{empty}
\pagestyle{empty}

%%%%%%%%%%%%%%%%%%%%%%%%%%%%%%%%%%%%%%%%%%%%%%%%%%%%%%%%%%%%%%%%%%%%%%%%%%%%%%%%
\begin{abstract}
Remote access to robotics hardware does not by itself guarantee usable experimental data. A teleoperated trial can complete the task and still produce an incomplete or inconsistent record. PUPPET aims to solve this problem by presenting a cyber-physical framework that opens a physical robotics laboratory through a continuously synchronized VR replica of the robot, stranding environment, and task objects. In addition, PUPPET integrates a trial validation pipeline built into the experimental workflow. We evaluated PUPPET in a within-subject study with 30 participants, who performed the same table-to-shelf object-transfer task in person, through 2D teleoperation, and through PUPPET. VR reduced post-learning completion time by 27.7\% over 2D (68.7\,s vs.\ 95.0\,s), statistically indistinguishable from in-person execution (72.1\,s). It also produced more validation-ready (85.6\% vs.\ 66.0\%) and intervention-free (89.7\% vs.\ 67.2\%) trials. With PUPPET we show that modality shapes the remote task performance and the reusability of the resulting experimental record.
\end{abstract}

%%%%%%%%%%%%%%%%%%%%%%%%%%%%%%%%%%%%%%%%%%%%%%%%%%%%%%%%%%%%%%%%%%%%%%%%%%%%%%%%

\section{INTRODUCTION}
Recent advances in physical AI increasingly depend on real-world interaction and the collection of physical experimental data~\cite{gemini_robotics_2025}. As robotic systems expand across healthcare, industrial automation, logistics, domestic assistance, and collaborative applications \cite{vanderloos2020service,Badue2021}, access to the underlying infrastructure remains a major bottleneck. Robots and laboratory setups are expensive, geographically fixed, and difficult to replicate. Conducting and repeating real-robot experiments is therefore still strongly tied to physical presence in a laboratory.

Remote laboratories, digital twins, and immersive teleoperation systems  have reduced this barrier by making physical robots accessible over distance~\cite{jara2009remote,pickem2017robotarium,zahid2024cloudgripper,jones2020digital,wonsick2020vr,wang2024xr}. These platforms are mainly designed for algorithmic execution, benchmarking, education, or constrained remote interaction.~\cite{pickem2017robotarium,zahid2024cloudgripper,xu2021digital}. 
Conventional 2D teleoperation provides remote access but limits spatial awareness, requiring users to reconstruct the workspace from separate camera views. More recently, VR and XR systems have enabled more immersive robot control and large-scale demonstration collection ~\cite{zhang2018deep,whitney2018ros,george2025openvr,moletta2023virtual}. Despite these advances, existing platforms are still primarily evaluated in terms of task performance, interface usability, or control quality. What remains unexplored is whether the mode of access to a physical laboratory also affects the quality and usability of the experimental data produced.

Our work addresses several questions: Can immersive VR serve as an access layer to a physical robotics laboratory to record experimental data that is complete, consistent, and supports reuse? Does an embodied virtual environment change how users experience and manage a real-robot experiment?

In this work, we present PUPPET (Physical-Unified Platform for Presence, Embodiment, and Teleoperation), a cyber-physical framework designed around the observation that successful teleoperation is not necessarily the same as successful experimentation. For instance, a remotely controlled trial may complete the task but still produce an incomplete, inconsistent, or unusable data. PUPPET addresses this by providing access to a physical robotics laboratory through a synchronized VR environment that integrates the robot, task objects, workspace, and experimental context within the same virtual space. Data collection and validation are built into the workflow, so that each trial can be assessed by task outcome and whether the resulting data is complete, consistent, and supports reuse.

\begin{figure}[t]
    \centering
    \includegraphics[width=0.9\columnwidth]{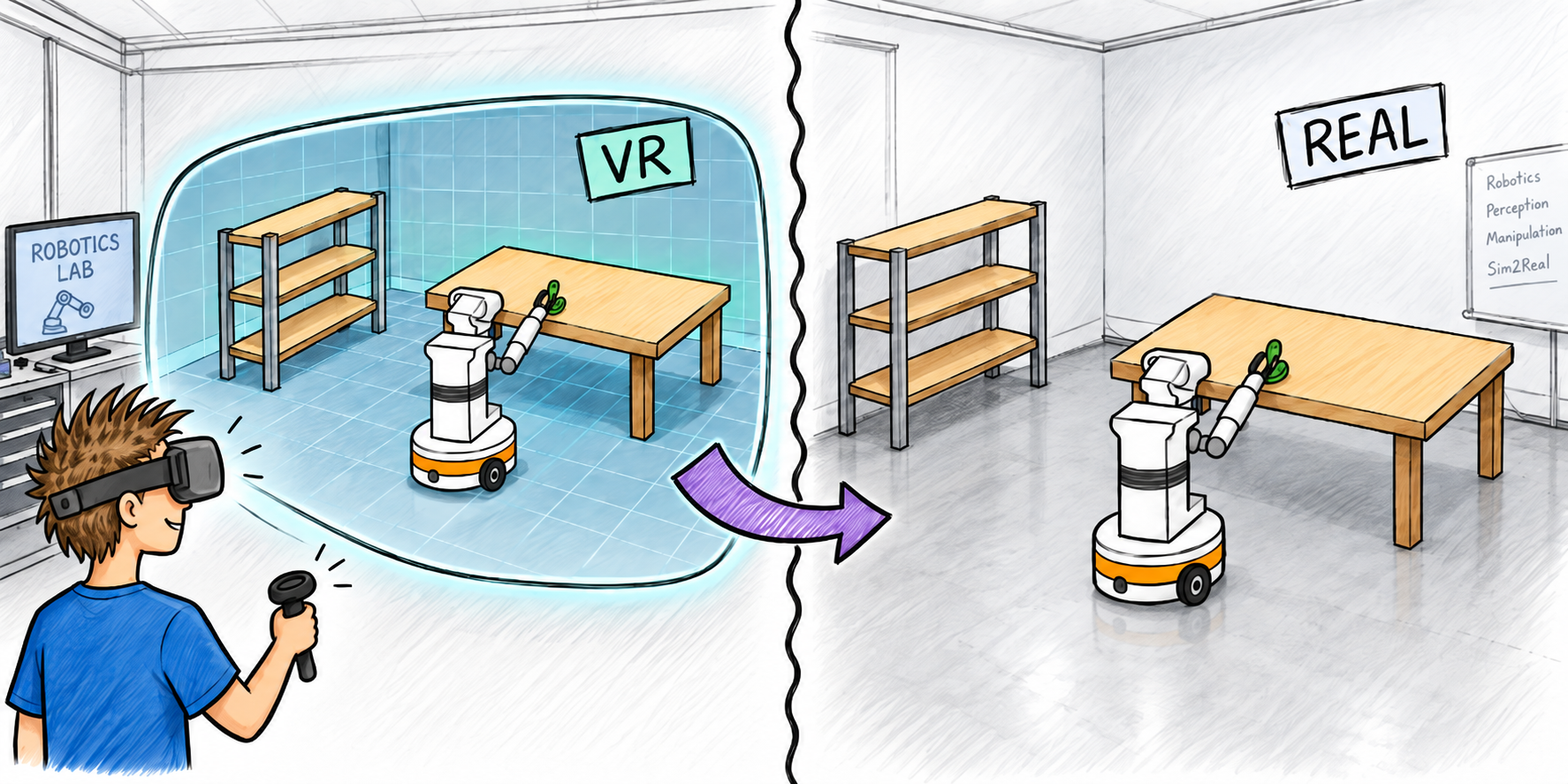}
    \caption{PUPPET provides an immersive virtual
    representation of the physical robotics laboratory,
    allowing users to interact with the real robot through
    the replicated VR environment.}
    \label{fig:puppet_sketch}
    \vspace{-5mm}
\end{figure}

We evaluated PUPPET through a within-subject user study with 30 participants, in which each participant performed the same mobile manipulation task on a PAL Robotics TIAGo robot~\cite{paltiagohardware} under three conditions: in-person, conventional 2D, and VR-based access. By keeping the robot, task, and physical setup fixed across conditions, the study isolates the effect of the access modality itself. Each condition includes repeated trials, allowing us to examine both task execution and learning across attempts. The evaluation addresses task effectiveness, reliability of the experimental record, and user experience.

Our results show that, after the initial learning trial, VR reduced mean completion time by 27.7\% relative to 2D teleoperation (68.66~s vs.\ 94.96~s), while remaining similar to the in-person execution time (72.10 s) and maintaining the same task success. For experimental reliability, VR produced more intervention-free trials than 2D (89.7\% vs.\ 67.2\%) and more validation-ready trials (85.6\% vs.\ 66.0\%). For user experience, VR significantly reduced workload (2.26 vs.\ 2.81), increased SUS usability scores (3.78 vs.\ 2.43) and increased task confidence (4.43 vs.\ 3.50) relative to 2D teleoperation.
In summary, the contributions of our work are:

\begin{itemize}
    \item A cyber-physical virtual laboratory that provides immersive access to a real robotics experiment by synchronizing the robot, workspace, and task-relevant objects in a continuously updated VR environment.
    \item An integrated validation pipeline that assesses whether each remotely conducted trial produces a complete, consistent, and reusable experimental record.
    \item A controlled within-subject study comparing in-person, 2D, and VR-based access across repeated trials, evaluating task effectiveness, experimental reliability, and user experience.
\end{itemize}

\section{Related Work}
\subsection{Remote and Shared Robotic Laboratories}
Remote and shared robotic laboratories have been proposed to extend access to robotic hardware, as real robots are expensive, difficult to maintain, and not equally available to all researchers. The Robotarium is a representative example, providing remote access to a swarm robotics testbed~\cite{pickem2017robotarium}. More recent platforms such as CloudGripper extend remote access toward manipulation research by exposing a set of robotic arms for benchmarking, experimentation, and large-scale data collection~\cite{zahid2024cloudgripper}. Related remote robotics laboratories have also been widely used in education, where students and users interact with real robots through web interfaces, video streams, or APIs~\cite{candelas2003teaching, theconstructRemoteLab}. These systems demonstrate the value of making physical robots remotely accessible, but they are primarily designed for algorithm execution, benchmarking, or structured educational interaction. 

\subsection{Digital Twins and Immersive Teleoperation}
Digital twins have become an important paradigm for pairing physical robots with virtual representations that support remote operation and human-in-the-loop interaction~\cite{xu2021digital,li2022ar}. Recent systems have demonstrated this across several configurations: modular plug-and-play frameworks for robotic-arm teleoperation~\cite{audonnet2024telesim}, mixed-reality extensions that bridge user and robot inside an immersive environment~\cite{audonnet2024immertwin}, AR-based synchronized visualization for collaborative manufacturing~\cite{xu2021digital,li2022ar}, and XR telepresence for shared remote robot programming~\cite{calandra2024digital}.

In parallel, VR has emerged as a practical interface for robot teleoperation, with prior work demonstrating remote robot control and demonstration collection for imitation learning~\cite{zhang2018deep} and direct interaction with ROS-enabled robots~\cite{whitney2018ros}. More recent works have explored VR for industrial manipulators, shared-control manipulation, and human-robot collaboration tasks~\cite{rosen2023virtual,xu2022shared,moletta2023virtual}. Frameworks such as OpenVR and LeVR further integrate VR teleoperation with imitation-learning pipelines, enabling scalable demonstration collection~\cite{george2025openvr,weng2025levr}. 

Despite these advances, both digital-twin and VR teleoperation systems remain largely focused on robot control, task-specific interaction, or demonstration collection~\cite{zhang2018deep,whitney2018ros,rosen2023virtual,george2025openvr,audonnet2024telesim,audonnet2024immertwin}. In real robotics experiments, however, the robot trajectory is often only part of what matters. The state of the surrounding environment can also be part of the experimental data. For example, when collecting demonstrations for a cloth-folding policy, the robot motion alone does not fully describe the demonstration; the cloth configuration, its position on the table, and the spatial context of the manipulation also matter. The same is true for experiments that depend on objects, fixtures, obstacles, or other changing parts of the workspace. A virtual interface centered only on the robot and its immediate target can therefore support teleoperation but still lose part of the context needed to understand or reuse the resulting data. What appears comparatively underexplored is integrating embodied access to the physical laboratory with specific validation of the final experimental record. PUPPET addresses this gap by treating the synchronized laboratory environment and the resulting experimental record as part of the same system, combining embodied interaction with validation of whether each real-robot trial is complete, consistent, and suitable for later analysis or reuse.
\section{System Design}
\label{sec:system}
\begin{figure*}[t]
    \centering

    \vspace*{3mm} % moves figure slightly down from top margin

    \begin{minipage}{0.88\textwidth}
        \centering

        \includegraphics[
            width=\linewidth
        ]{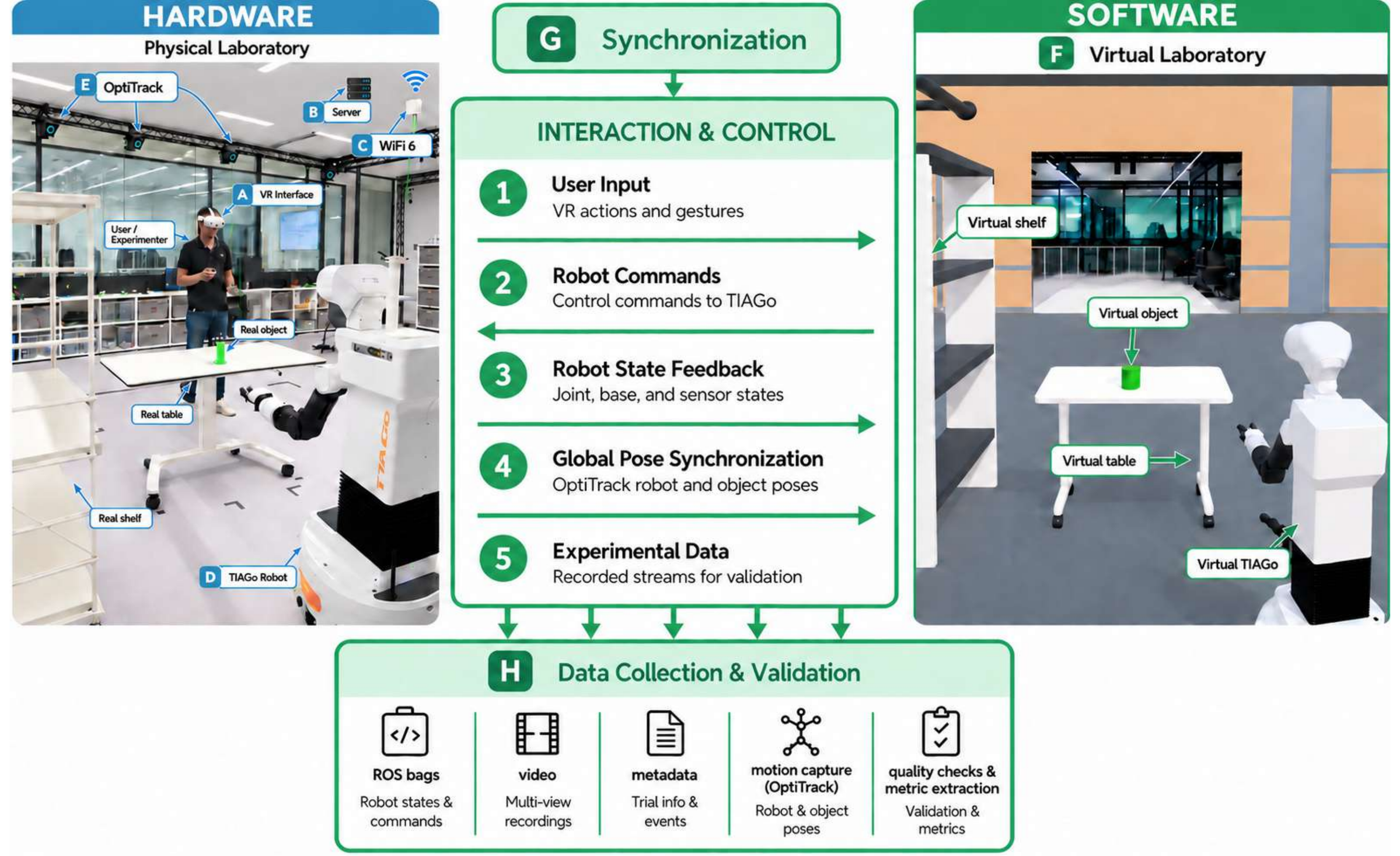}

        \vspace{-1mm}

        \caption{
        The physical and virtual laboratories are connected through interaction and control, synchronization, and an integrated data collection and validation layer. Letters A--I correspond to the system components described in the text, and numbered arrows indicate the main control and data flows.
        }
        \label{fig:puppet_architecture}
    \end{minipage}

    \vspace{-2mm}
\end{figure*}

PUPPET is a cyber-physical robotics testbed that provides remote access to a physical laboratory through an immersive VR interface. In the use case presented in this work, a remote user enters a virtual replica of the laboratory, controls the physical TIAGo robot through its virtual counterpart, and performs an object-transfer task executed in the real workspace\footnote{The source code and data will be released upon acceptance.}. The architecture in Fig.~\ref{fig:puppet_architecture} integrates the virtual laboratory, robot control, and data validation in a single workflow.

\subsection{Physical Laboratory and Synchronization}

Remote researchers enter the testbed through the VR interface (A), a Meta Quest~3 headset with haptic controller straps that provide vibro-tactile feedback and joystick input. The headset tracks the user's head with 6 DOF, the straps track both hands, and the controllers register joystick and button input. These tracked poses and button events are streamed to the server (B), which renders the virtual laboratory, hosts the robot control interfaces, and closes the control loop at 100\,Hz. From there the commands travel over a commercial WiFi~6 access point (C) to reach the physical robot (D) that is in the laboratory. The measured VR command-to-motion feedback latency had a median of 30.6 ms.

The robot used in this work is TIAGo~\cite{paltiagohardware} (D), which combines a differential-drive mobile base with a 7-DOF torque-controlled arm~\cite{paltiagohardware}. It is further equipped with a head-mounted RGB-D camera, joint encoders providing proprioceptive feedback, and an onboard PC running its ROS control stack. TIAGo was selected because it couples mobile navigation with arm manipulation, so a single task exercises the full teleoperation pipeline and produces rich, multi-modal experimental data. 

Independent of the robot's own position estimates, an OptiTrack motion-capture system (E) tracks the laboratory at 120 Hz with sub-millimeter accuracy. Rigid-body markers on the mobile base and on task-relevant objects provide the 6-DOF poses used both for physical--virtual alignment and as the recorded object trajectories. 

The physical and virtual laboratories are held together by the synchronization layer (G). The NatNet ROS driver streams the OptiTrack poses into the ROS network, and the ROS--TCP connector translates them into the transforms that position each object in the virtual scene in real time. End-to-end delay from physical motion to virtual update was 7.27 ms, with a median of 7.15 ms.

\subsection{Virtual Laboratory and Embodied Control}
The virtual laboratory (F) is built in Unity 2022.6.3f1 with the Meta VR SDK. It reproduces the full physical laboratory including walls, floor, tables, shelves, TIAGo, and task-relevant object. The user sees the entire experimental workspace that serves as  primary teleoperation interface and a live visualization for robot state, sensor data, and task progress.

Within that virtual laboratory, interaction and control are embodied. The user's arm becomes the TIAGo arm and the user's head becomes the TIAGo head. The tracked hand pose drives the end-effector directly, with motion planning enforcing the arm's kinematic limits. The gripper is actuated by discrete open and close commands. Headset orientation drives the TIAGo head, so the viewpoint follows the user's gaze and the user can look in one direction while reaching in another. The robot base is driven by the controller joystick.

\subsection{Data Collection and Validation}
The streams that keep the virtual laboratory synchronized are also written to disk as the experimental record (H). Each trial stores a ROS bag with joint states, commanded velocities, base odometry, gripper state, end-effector pose, and the OptiTrack poses of the robot and task object, together with trial metadata and multi-view video at 30 fps. Data collection runs on the robot side and does not depend on how the workspace is accessed. The data validation runs automatically when a trial closes and emits a per-trial report. 

No established standard exists for validating teleoperated robotics data, so we operationalize four quality dimensions drawn from prior work on teleoperation and demonstration quality~\cite{narayanan2026dqaf,sakr2025consistency,yu2026warprm}. These quality dimensions are completeness, temporal continuity, signal plausibility, and execution consistency. These four quality dimensions are operationalized through the six rule families listed in Table~\ref{tab:checks}. Each dimension becomes one or more specific validation rules: completeness is evaluated through R1, temporal continuity through R2, signal plausibility through R3--R4, and execution integrity and consistency through R5--R6. R1 is further divided into record completeness (R1a) and validation metadata (R1b). Each rule returns Fail when the record is unrecoverable, Review when it is complete but anomalous, or Pass when no quality issues are detected.

The rules apply criteria that are meant for the present use case. These criteria were fixed to mark points when a data signal became either practically unusable or inconsistent with the expected behavior of the task. Under R2 (telemetry continuity), end-effector data are reviewed when more than 90\% of samples are missing, as the remaining data is too sparse to reconstruct meaningful motion. Similarly, required telemetry streams are also reviewed when more than 50\% of samples are missing, and a timestamp gap above 1\,s is treated as a discontinuity. Under R3 (tracking plausibility), trials are reviewed when the object path is below 0.02\,m or when tracked-object movement is below 0.005\,m on all axes. R4--R6 extend these checks to robot motion, technician intervention, and consistency across the repeated trials.

\begin{table*}[t]
\centering

\vspace*{5mm} % move entire table down from top margin

\captionsetup{width=0.84\textwidth}
\caption{Validation rules implementing the four data-quality dimensions used in the study.}
\label{tab:checks}

\footnotesize
\setlength{\tabcolsep}{2.5pt}
\renewcommand{\arraystretch}{1.12}

\begin{adjustbox}{max width=0.90\textwidth,center}
\begin{tabularx}{0.90\textwidth}{@{}
>{\raggedright\arraybackslash}p{0.04\textwidth}
>{\raggedright\arraybackslash}p{0.14\textwidth}
>{\raggedright\arraybackslash}p{0.14\textwidth}
>{\raggedright\arraybackslash}X
>{\centering\arraybackslash}p{0.055\textwidth}
@{}}
\toprule
\textbf{Rule} &
\textbf{Quality dimension} &
\textbf{Check} &
\textbf{Criteria} &
\textbf{Outcome} \\
\midrule

R1a &
Completeness &
Record completeness &
Required metadata, files, topics, telemetry columns, and minimum sample count must be present. &
Fail \\[2pt]

R1b &
Completeness &
Validation metadata &
Missing manual- or video-validation information is flagged for inspection. &
Review \\[2pt]

R2 &
Temporal continuity &
Telemetry continuity &
Review if required streams are $>50\%$ missing, end-effector pose is $>90\%$ missing, or the largest timestamp gap exceeds 1\,s. &
Review \\[2pt]

R3 &
Signal plausibility &
Object tracking &
Review successful trials if object path is $<0.02$\,m or movement is $<0.005$\,m on all axes. &
Review \\[2pt]

R4 &
Signal plausibility &
Robot motion &
Review if expected gripper, arm, torso, or end-effector motion is too small, or joint jerk exceeds \(5\times10^{4}\). &
Review \\[2pt]

R5 &
Execution integrity and consistency &
Technician intervention &
Review if there are more than two minor interventions, any major intervention in T2--T3, or inconsistent intervention metadata. &
Review \\[2pt]

R6 &
Execution integrity and consistency &
Cross-trial consistency &
Review when repeated-trial variation in completion time, path length, gripper chatter, or joint jerk exceeds predefined limits. &
Review \\

\bottomrule
\end{tabularx}
\end{adjustbox}

\end{table*}

\section{User Study}
\label{sec:user_study}

This study evaluates whether immersive VR can serve as an access layer to a physical robotics laboratory that produces reusable experimental records, and whether an embodied virtual environment changes how users experience and manage a real-robot experiment. To address these questions, we conducted a controlled user study comparing three modes of access to the same robot and task: in-person operation, conventional 2D remote operation, and PUPPET. The evaluation was conducted using a robotic object-transfer task performed with the TIAGo robot, and is structured around the three hypotheses defined below.

\subsection{Hypotheses}
\textbf{H1}: PUPPET will support more effective remote task execution than conventional 2D teleoperation, reflected in comparable task completion and lower post-learning completion time.

\textbf{H2}: PUPPET will improve the reliability of remote experimentation compared with conventional 2D teleoperation, reflected in a greater proportion of validation-ready trial records, fewer technician interventions, and maintained consistency across repeated executions.

\textbf{H3}: Users will report a more favorable interaction experience in VR than with conventional 2D teleoperation, reflected in workload, usability, and task confidence.

\subsection{Method}
We used a within-subject experimental design in which each participant completed the same robotic manipulation task under three conditions presented in counterbalanced order: in-person operation, 2D camera-based remote operation, and VR-based remote operation through PUPPET. The task required participants to use the TIAGo mobile manipulator to pick up a plastic cylinder from a table and place it on a shelf. At the start of each trial, the robot was positioned at a home pose 1.7m from the table. A trial was considered complete when the object was placed on the shelf. Between conditions, participants took a 5-minute break to reduce fatigue and carry-over effects.

Before the formal trials, participants received condition-specific training consisting of a guided walkthrough of the interface and one practice trial to get comfortable with the control method. Training duration was approximately 10 minutes per condition to ensure comparable familiarization across modalities. Participants also completed a brief questionnaire on prior technology experience and demographics.

The three conditions shared the same robot, task, and underlying ROS control pipeline, differing only in the interface through which participants accessed the workspace: \emph{i)} \textit{In-person}: participants operated the TIAGo while physically present in the laboratory, with direct visual access to the robot and task environment. Robot remote control was provided via a keyboard for the robot arm control and joystick for the mobile base; \emph{ii)} \textit{2D}: participants performed the task remotely using 3 camera feeds displayed on a single monitor. Robot remote control was provided through a keyboard for the robot arm control and joystick for the mobile base;; and \emph{iii)} \textit{VR}: participants performed the task remotely through PUPPET, using the Meta Quest 3 headset and haptic controller straps to interact with a synchronized virtual representation of the physical laboratory.

After completing the background questionnaire, participants performed the task under each of the three conditions and completed a post-condition questionnaire after each trial. Condition order was counterbalanced using a balanced Latin square across six orderings to control for learning and fatigue effects. Within each condition, participants completed three repetitions of the same task. The first trial (T1) served as the learning trial; second (T2) and third (T3) trials were treated as post-learning trials for the primary analysis, following standard practice in teleoperation studies where initial exposure produces the largest performance shift.

\subsection{Measures}
The study evaluated the three interaction conditions using objective task-performance measures, subjective user-experience measures, and system-level data-quality measures. Task performance was assessed using task completion rate, task completion time, and technician interventions. A trial was considered complete when the participant completed the full table-to-shelf transfer without a safety stop. Technician assistance was recorded separately as an intervention measure. Completion time was measured from trial initiation (first robot command) until successful placement of the object at the goal. Failed trials were excluded from the completion-time analysis but included in the completion rate and intervention counts.

After each condition, participants completed a questionnaire. Subjective workload was assessed using the NASA Task Load Index (NASA-TLX)~\cite{hart1988development}, rated on a 1--5 scale, with higher scores indicating greater workload. System usability was measured using the System Usability Scale (SUS)~\cite{brooke1996sus}. Responses were converted to the standard 0--100 SUS score and subsequently normalized to a 0--5 scale for comparison and analysis alongside the other questionnaire measures.Task confidence was rated on a 1--5 scale, with higher scores indicating greater confidence in repeating the task independently. Sense of agency, defined as the user's subjective experience that their own actions cause the resulting robot motion and behavior, was assessed only after the VR condition. Three 1--5 items measured perceived control over the robot, correspondence between participant actions and robot motion, and whether the robot responded as expected; higher scores indicated greater perceived agency.

Data quality was evaluated using the PUPPET validation pipeline described in Sec.~\ref{sec:system}. Each trial was classified as Pass (all automated checks satisfied), Review (minor issues flagged for inspection), or Fail (critical data missing or corrupted), then characterized using the execution metrics extracted from the recorded telemetry: joint-space path length, joint jerk, joint effort, base movement, gripper behavior, and object movement. Cross-trial consistency was assessed by computing the coefficient of variation (CV) of these execution metrics across T2–T3 within each condition, following consistency-based demonstration-quality approaches~\cite{sakr2025consistency}. Lower variability across repeated trials indicates more consistent execution and more predictable experimental data. The resulting measures were used to compare the completeness, execution characteristics, and consistency of the data collected across in-person, 2D, and VR interaction.

Statistical methods to analyze were chosen based off of the repeated-measures structure of the study. The overall differences across the three conditions were evaluated using Friedman tests, with planned pairwise comparisons between VR and 2D conducted using two-sided Wilcoxon signed-rank tests. The same paired method was used for participant-level reliability measures and questionnaire outcomes, with rank-biserial correlations reported as effect sizes. Sense of agency in VR was additionally tested against the neutral midpoint using a one-sample Wilcoxon signed-rank test. Statistical significance was assessed at $\alpha=.05$, with exact $p$-values and effect sizes reported alongside descriptive statistics.

\subsection{Participants}
We recruited N = 30 adult participants from STEM backgrounds with different levels of prior experience with robotics, VR, and remote-control interfaces. Prior experience with TIAGo or PUPPET was not required, as all participants received the same training before starting the experimental trials. 

\section{Results}
\begin{figure}[tbh]
    \centering
    \includegraphics[width=0.90\columnwidth]{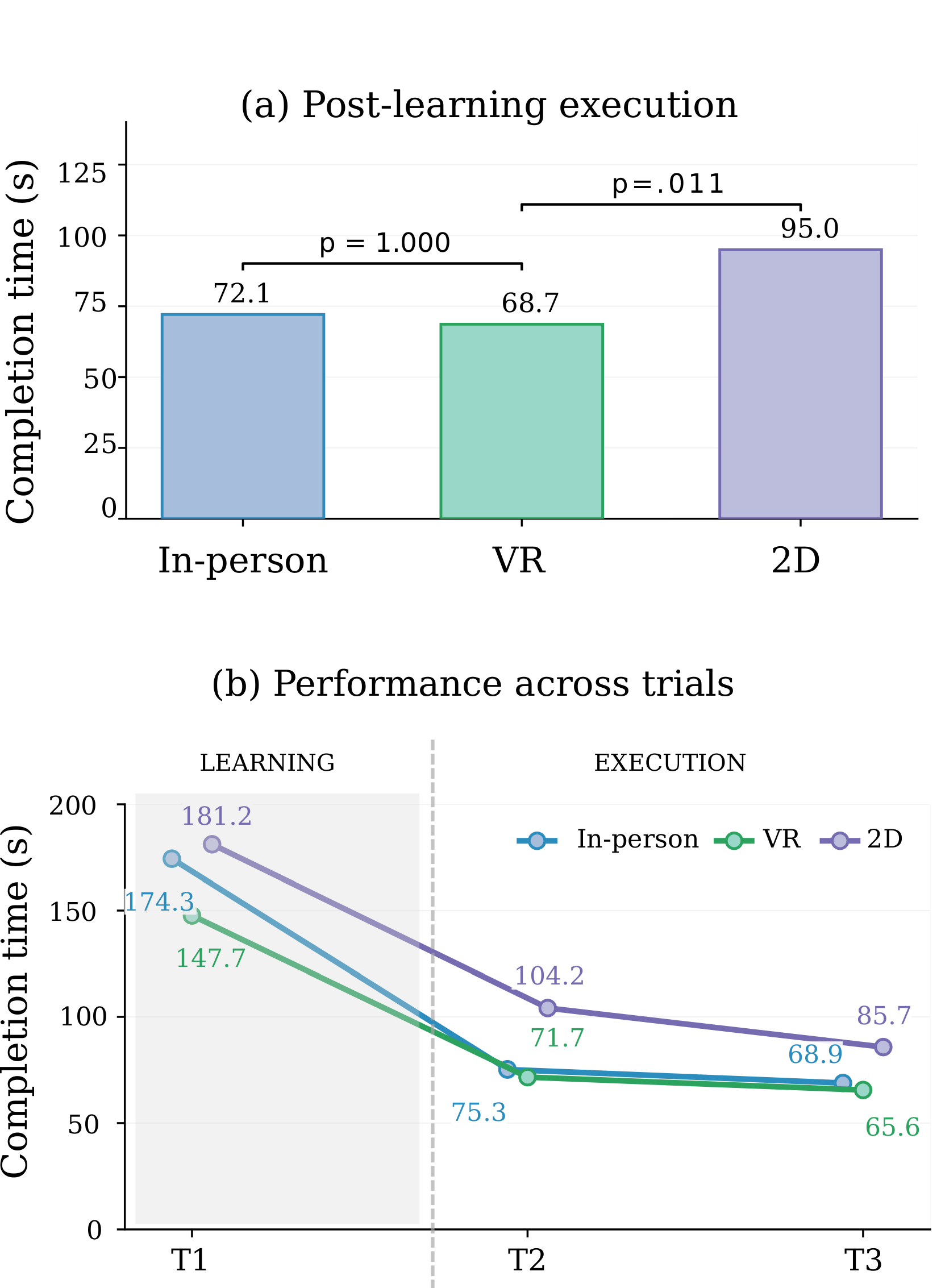}
    \caption{H1 task performance: (a) post-learning execution time and (b) completion time across repeated trials.}
    \label{fig:h1_completion_time}
    \vspace{-2mm}
\end{figure}

\subsection{H1: Task Effectiveness}

Fig.~\ref{fig:h1_completion_time} compares task completion time across the three interaction conditions.  The top graph (Fig.~\ref{fig:h1_completion_time} a) reports post-learning execution time. Participants completed the task in 68.66 s through PUPPET, in 94.96 s through 2D teleoperation, and in 72.10 s in person, a 27.7\% reduction from 2D to VR. Twenty of the 30 participants were faster in VR than in 2D, and the difference was significant ($W = 109.00$, $p = .011$, $r = -.531$). Completion time also differed across the three conditions overall ($\chi^2(2) = 6.10$, $p = .047$). Condition order had no detectable effect on completion time (\(p=.639\)).

Fig.~\ref{fig:h1_completion_time}(b) shows how completion time changed across repetitions. On T1 participants were faster through PUPPET (147.7\,s) than through 2D (181.2\,s) or in person (174.3\,s). All conditions then improved sharply. Between T2 and T3, the VR times were stable, falling from 71.7 s to 65.6 s, while 2D was still improving from 104.2 s to 85.7 s. H1 was overall supported.
\subsection{H2: Reliability of Remote Experimentation}
\begin{figure*}[t]
    \centering

    \vspace*{5mm} % moves figure down from top margin

    \includegraphics[
        width=0.94\textwidth
    ]{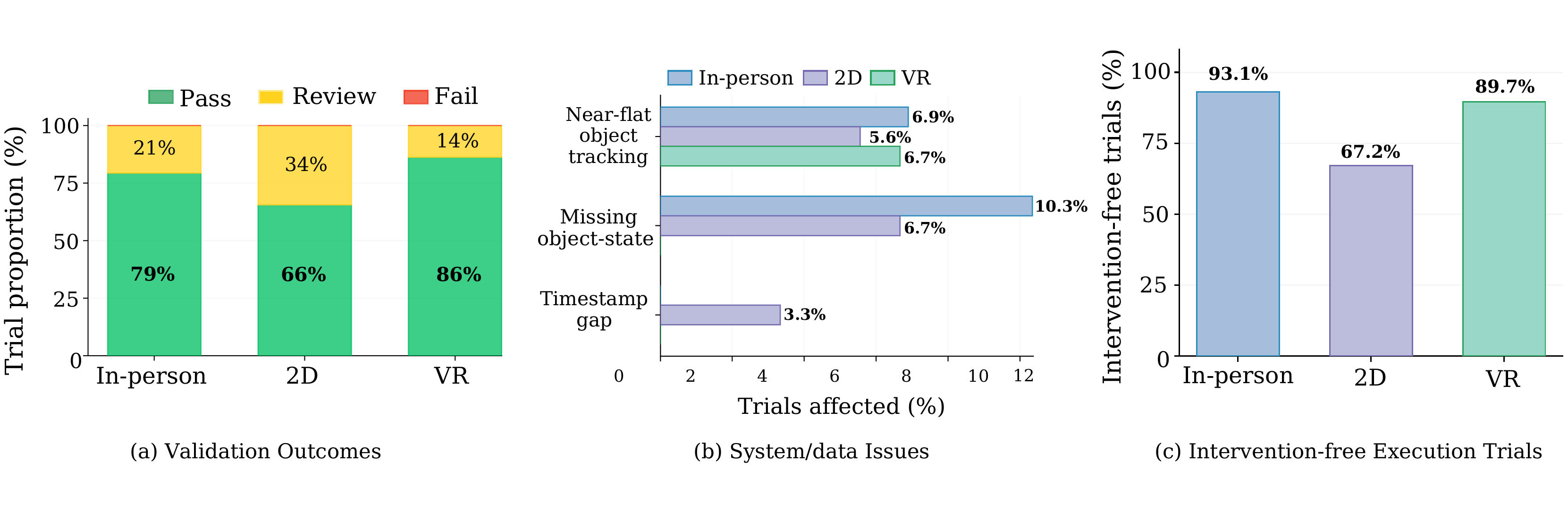}

    \vspace{-2mm}

    \caption{
    H2 reliability outcomes. Issues in (b) are non-exclusive and only top categories selected.
    }
    \label{fig:h2_results}

    \vspace{-4mm}
\end{figure*}

\begin{figure}[t]
    \centering

    \includegraphics[
        width=0.93\columnwidth
    ]{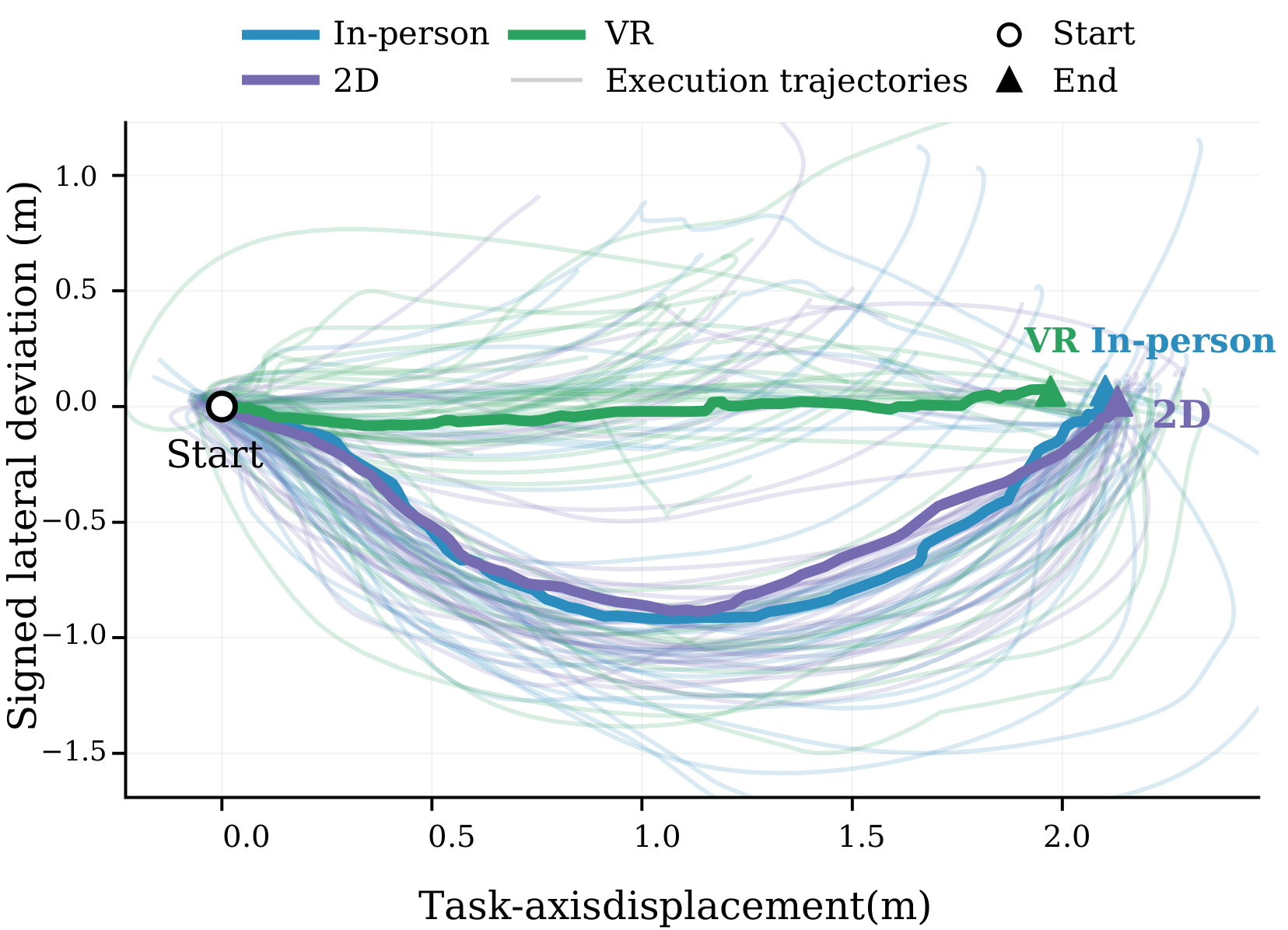}

    \vspace{-1mm}

    \caption{
    Task-aligned object trajectories during execution trials (T2--T3). Light lines show individual executions, while bold lines show the median trajectory for each condition.
    }
    \label{fig:h2_repeatability}

    \vspace{-3mm}
\end{figure}

Fig.~\ref{fig:h2_results} summarizes experimental reliability. As shown in Fig.~\ref{fig:h2_results}(a), 85.6\% of VR trials were validation-ready against 66.0\% of 2D trials, and the difference was significant at the participant level ($W = 21.00$, $p = .012$, $r = .691$). No trial in any condition was assigned Fail. Every flagged trial fell into Review, which means the failures we observed were partial or implausible data rather than unrecoverable loss.

The issue profile in Fig.~\ref{fig:h2_results}(b) shows the main reasons for a trial to be flagged for review. Near-flat object tracking affected 6.9\% of in-person trials, 5.6\% of 2D trials, and 6.7\% of VR trials. Missing object-state data affected 10.3\% of in-person trials and 6.7\% of 2D trials and did not
occur in VR at all. Large timestamp gaps appeared only in 2D (3.3\%). Fig.~\ref{fig:h2_results}(c) reports intervention-free execution over the post-learning trials. 89.7\% of VR trials ran without technician assistance, against 67.2\% in 2D and 93.1\% in person, and VR required significantly fewer interventions than 2D ($W = 24.50$, $p = .007$, $r = -.680$).

Figure~\ref{fig:h2_repeatability} shows task-aligned object trajectories for T2 and T3. The VR median path was markedly more direct. Median peak lateral deviation was 0.713\,m in VR, 0.925\,m in 2D and 1.028\,m in-person, although the overall condition effect did not reach statistical significance (\(\chi^2(2)=5.15,\ p=.076\)). Repeatability between T2 and T3 did not differ between VR and 2D for RMS distance ($p = .956$), dynamic time warping distance ($p = .956$), or path-length difference ($p = .312$). Embodied arm control therefore changed the shape of the motion without reducing consistency at the object level, though these comparisons show only that we did not detect a difference. H2 was supported.

\subsection{H3: User Experience}

\begin{figure}[t]
    \centering

    \vspace*{2mm} % moves figure slightly down from top margin

    \begin{minipage}{0.90\columnwidth}
        \centering

        \includegraphics[
            width=\linewidth
        ]{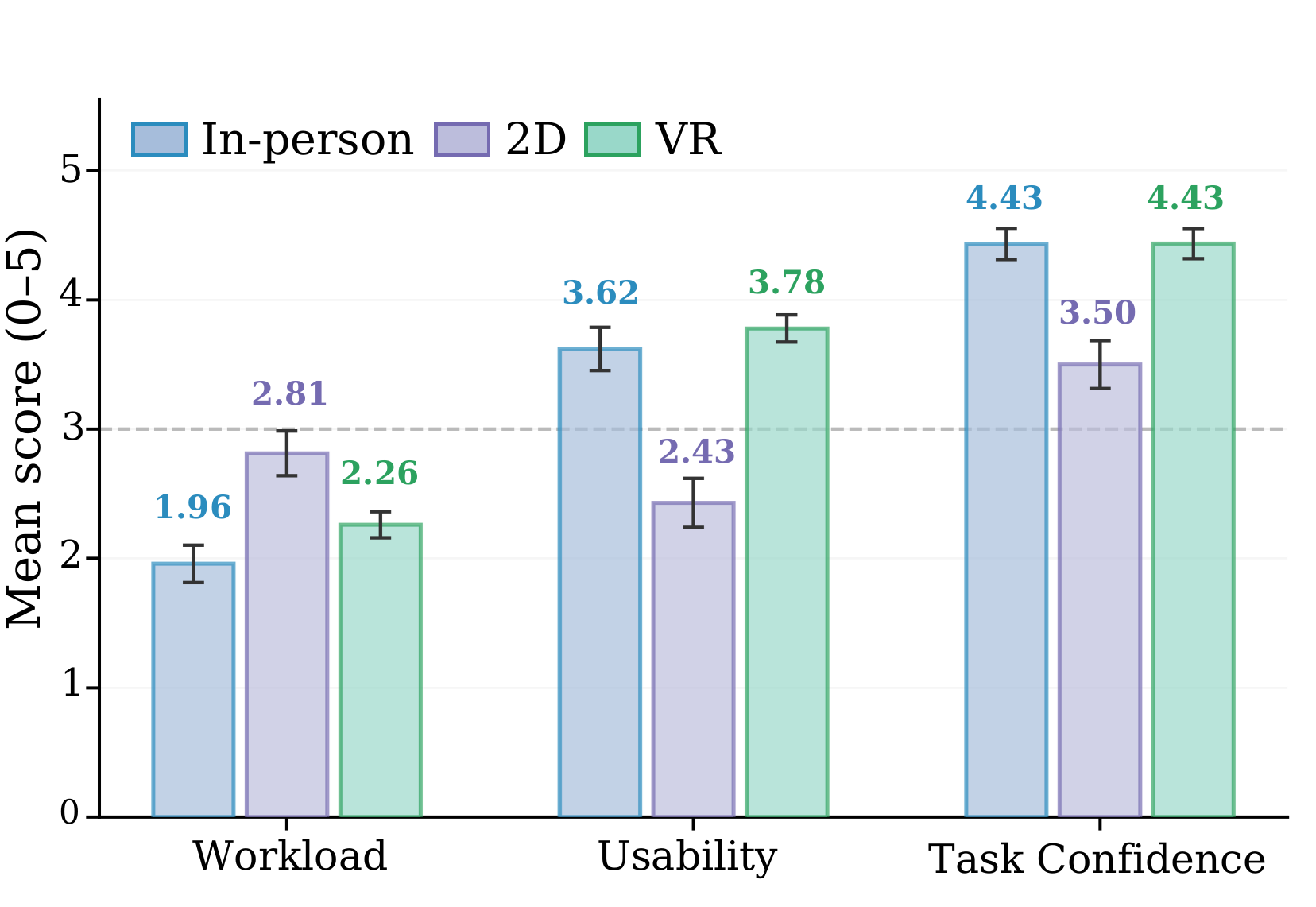}

        \vspace{-1mm}

        \caption{
        User-experience outcomes across conditions. SUS scores are normalized to a 0--5 scale for visualization; statistical analyses use standard 0--100 SUS scores.
        }
        \label{fig:h3_results}
    \end{minipage}

    \vspace{-3mm}
\end{figure}

\begin{table}[t]
\centering
\caption{Sense of agency reported in the VR condition. Scores range from 1 to 5, with higher values indicating greater perceived agency.}
\label{tab:vr_agency}
\small
\setlength{\tabcolsep}{4pt}
\begin{tabular}{lccc}
\hline
\textbf{Agency measure} & \textbf{Mean} & \textbf{SD} & \textbf{Median} \\
\hline
Perceived control over the robot & 4.33 & 0.55 & 4.00 \\
Robot motion reflected user actions & 4.27 & 0.69 & 4.00 \\
Robot responded as expected & 4.33 & 0.61 & 4.00 \\
\textbf{Overall agency} & \textbf{4.31} & \textbf{0.53} & \textbf{4.33} \\
\hline
\end{tabular}
\vspace{-1mm}
\end{table}

Figure~\ref{fig:h3_results} consolidates workload, usability, and task confidence across the three conditions on a common 0--5 scale. VR showed lower workload than 2D (2.26 vs.\ 2.81), higher usability (3.78 vs.\ 2.43), and higher task confidence (4.43 vs.\ 3.50), while in-person scores were 1.96, 3.62, and 4.43, respectively. In the paired VR--2D comparisons, 18 of 30 participants reported lower NASA-TLX workload in VR (2.26 vs.\ 2.81; $W=73.00$, $p=.009$, $r=-.584$). The task confidence was also significantly higher in VR (4.43 vs.\ 3.50; $W=31.50$, $p<.001$, $r=.806$), with 21 of 30 participants reporting higher confidence. Table~\ref{tab:vr_agency} shows high agency scores in VR, with an overall mean of 4.31 ($SD=0.53$, $Mdn=4.33$). The overall score was above the neutral midpoint of 3 ($W=0.00$, $p<.001$, $r=1.00$). Overall, H3 was supported.

\section{Discussion}

In this paper, we investigated whether the mode of access to a physical robotics laboratory affects how well a remote experiment is performed and how usable data it records. The clearest finding is that these two outcomes came apart. Every trial in every condition reached successful task completion, so task success alone could not distinguish the three access conditions at all. The experimental data show that trials conducted through PUPPET were validation-ready far more often than those conducted through conventional 2D teleoperation, and required technician intervention far less frequently. A remotely controlled trial can therefore succeed as a task and still fail as a recorded experiment. This suggests that the quality of the resulting data deserves to be treated as an outcome measure in its own right.

The issue profiling further separates what VR teleoperation environments can and cannot address. Near-flat object tracking occurred at comparable rates in all three conditions and therefore reflects a property of the physical sensing rather than of the access modality. Large timestamp gaps appeared only in 2D teleoperation, and missing object-state data, observed in both other conditions, did not occur in VR at all. We attribute this last result to the tracked state that is visualized continuously inside the virtual laboratory. When an object freezes, disappears, or no longer follows the physical scene, users see it during the trial. The virtual laboratory therefore serves as a continuous check on the experimental record. 

The repeated trials also show how quickly users become familiar with each access modality. All conditions improved sharply after the learning trial, but VR completion times were stable between T2 and T3, while 2D was still improving. If early trials must be discarded because they reflect learning, the share of a data collection session that produces usable records depends on how long users take to learn the interface. For large-scale demonstration collection, where user time is the limiting resource, this matters as much as the quality of each individual trial.

Finally, embodied control changed how users chose to perform the task without reducing the consistency with which they repeated it. The condition-level trajectories differed in shape, yet object-level repeatability between T2 and T3 did not differ between VR and 2D. This is useful when considering robotics datasets more broadly, since a system can allow different valid strategies while protecting repeatability within each strategy. The user-experience results point the same way, with VR producing lower workload than 2D despite being the more complex system, and task confidence in VR matching that reported for in-person operation. The cost of remote operation appears to come from having to reconstruct spatial relationships from a limited set of camera views, rather than
from system complexity.

\subsection{Limitations and future work}

Our study has several limitations. First, we compare complete interaction methods rather than separating  individual component. In the 2D condition, users interacted with the robot through a combination of keyboard and joystick based teleoperation, while PUPPET combined a synchronized 3D workspace with embodied arm control. This reflects the actual difference between the two access methods we wanted to study, but it also means that the contribution of spatial visualization and embodied control cannot be separated from one another. Future work could use matched control schemes to determine their individual effects.

Second, our evaluation used one robot, one structured table-to-shelf task, and one laboratory setup. This controlled design makes the conditions easier to compare, but limits how far the results can be generalized to more complex tasks, other robots, longer experiments, or geographically distributed operation. The virtual laboratory is also manually modeled and depends on external motion capture, which limits deployment to instrumented labs and made marker loss a key source of VR data issues (6.7\% tracking anomalies). Tools such as SceneSmith~\cite{pfaff2026scenesmith}, together with markerless pose estimation could reduce this setup effort. Future work can also include representing people within the virtual laboratory and studying how remote users interact with human activity during real experiments.

Finally, our current study demonstrates repeatability within the same experimental setup rather than full reproducibility across laboratories, robot platforms, sessions, or independent operators. Future evaluations should test whether PUPPET and its validation pipeline transfer across tasks, systems, and sites while maintaining consistent experimental records.

\section{Conclusions}
In this paper, we present PUPPET, a cyber-physical platform that opens a physical robotics laboratory to remote researchers. A remote user acts on the robot through a continuously synchronized virtual replica of the workspace, and an integrated pipeline validates the record each trial leaves behind. We use a TIAGo robot performing a table-to-shelf object-transfer task to illustrate how the mode of access shapes both the remote operation and the data it produces, in a within-subject study comparing in-person operation, conventional 2D teleoperation, and VR-based access through PUPPET across repeated trials. Task success alone did not separate the three conditions; the differences appeared in the record, with VR-based access producing more validation-ready and intervention-free trials. Users also reported lower workload and higher usability, task confidence, and sense of agency in VR than in 2D, and object-level repeatability showed that the freedom introduced by embodied arm control does not come at the cost of consistency. Future work will extend PUPPET toward markerless environment synchronization, representation of human presence in the virtual laboratory, and validation across distributed sites.

\section*{Acknowledgments}
AI was used during the preparation of this manuscript for grammar and language editing, literature research, code assistance, and data visualization/plot generation. All research decisions, analyses, and final content were reviewed and verified by the authors.

\bibliographystyle{IEEEtran}
\bibliography{references}

\end{document}